\pdfoutput=1
\documentclass[11pt]{article}

\usepackage[preprint]{acl}

\usepackage{times}
\usepackage{latexsym}

\usepackage[T1]{fontenc}
\usepackage[utf8]{inputenc}

\usepackage{microtype}

\usepackage{inconsolata}

\usepackage{graphicx}
\usepackage{algorithm}
\usepackage{algorithmic}
\usepackage{subcaption}
\usepackage{multirow}
\usepackage{amsmath}
\usepackage{amsfonts}

\usepackage{mathtools}

\title{CRPO: Confidence-Reward Driven Preference Optimization for Machine Translation}

\author{
 \textbf{Guofeng Cui\thanks{The work was carried out while the first author was an intern at Amazon Prime Video}\textsuperscript{1}},
 \textbf{Pichao Wang\textsuperscript{2}},
 \textbf{Yang Liu\textsuperscript{2}},
 \textbf{Zemian Ke\textsuperscript{2}},
 \textbf{Zhe Liu\textsuperscript{2}},
 \textbf{Vimal Bhat\textsuperscript{2}}
\\
\\
 \textsuperscript{1}Rutgers University,
 \textsuperscript{2}Amazon
\\
 \small{
   \href{mailto:gc669@cs.rutgers.edu}{gc669@cs.rutgers.edu}, \href{mailto:}{\{pichaowang,kezemian\}@gmail.com}, \href{mailto:}{\{yangnliu,zhuzliu,vimalb\}@amazon.com}
}
}

\begin{document}
\maketitle
\begin{abstract}
Large language models (LLMs) have shown great potential in natural language processing tasks, but their application to machine translation (MT) remains challenging due to pretraining on English-centric data and the complexity of reinforcement learning from human feedback (RLHF). Direct Preference Optimization (DPO) has emerged as a simpler and more efficient alternative, but its performance depends heavily on the quality of preference data. To address this, we propose Confidence-Reward driven Preference Optimization (CRPO), a novel method that combines reward scores with model confidence to improve data selection for fine-tuning. CRPO selects challenging sentence pairs where the model is uncertain or underperforms, leading to more effective learning. While primarily designed for LLMs, CRPO also generalizes to encoder-decoder models like NLLB, demonstrating its versatility. Empirical results show that CRPO outperforms existing methods such as RS-DPO, RSO and MBR score in both translation accuracy and data efficiency.
\end{abstract}

\section{Introduction}

Recent advances in decoder-only large language models (LLMs), such as GPT series~\cite{achiam2023gpt}, LLaMA~\cite{touvron2023llama, dubey2024llama}, and Falcon~\cite{almazrouei2023falcon}, have showcased their outstanding ability to understand context and perform various natural language processing (NLP) tasks. However, applying LLMs to machine translation (MT) remains a challenging endeavor, especially due to their pretraining on predominantly English-centric datasets. This limitation has generated significant interest in aligning LLMs for translation tasks using further training methods, with particular attention to enhancing their multilingual performance.

\begin{figure*}
    \centering
    \includegraphics[width=\linewidth]{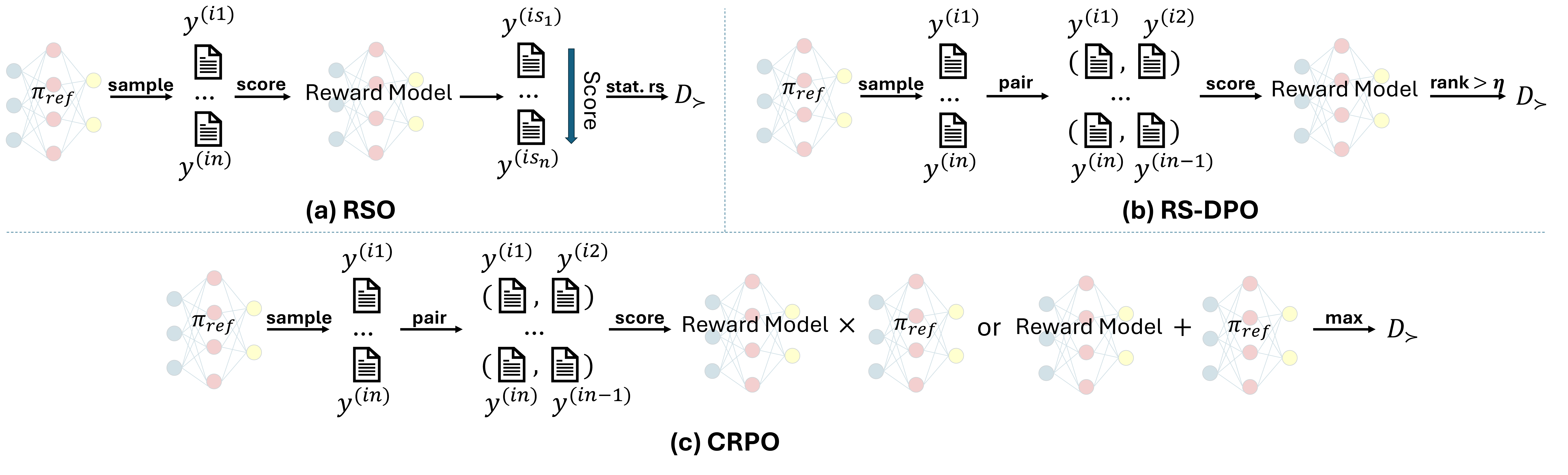}
    \caption{Comparison among RSO, RS-DPO and CRPO. RSO scores candidate responses with reward and applies statistical reject sampling for preference dataset. RS-DPO accepts sentence pairs with reward difference surpass a preset threshold. Instead, proposed CRPO evaluates sentence pairs with both reward different and policy confidence.}
    % \caption{Comparison among RSO, RS-DPO and CRPO. RSO samples candidate responses from SFT policy and then score each sentence with a trained pairwise reward model. Statistical rejection sampling is applied to subsample sentence for preference dataset, where the acceptance rate of each sentence is based on related reward. RS-DPO calculates reward difference of each response pair with point-wise reward model instead. Then it accepts the responses with reward different larger than a preset threshold. Proposed CRPO also calculates reward difference for all response pairs with point-wise reward model. But CRPO evaluates sentence pairs with both reward different and pretrained policy confidence.}
    \label{fig:comp}
\end{figure*}

To mitigate the linguistic bias inherent in LLMs, instruction tuning has become a widely adopted approach. Instruction tuning fine-tunes LLMs using multilingual datasets and translation-specific instructions, with the goal of expanding linguistic diversity and improving translation quality~\cite{yang2023bigtranslate, chen2023improving, zhu2023extrapolating, zhang2023bayling}. Despite these efforts, gaps remain between the performance of LLMs and specialized machine translation models~\cite{zhu2023multilingual}. To address these challenges, approaches such as reinforcement learning from human feedback (RLHF)~\cite{christiano2017deep} have been explored. RLHF allows LLMs to align with human preferences by training a reward model on human-annotated preference data and fine-tuning the LLM to maximize the predicted reward for translation quality. For example, \citet{xu2024advancing} construct a preference translation dataset using multilingual books and fine-tune LLaMA-2 with RLHF to optimize translation performance.

However, RLHF introduces several complexities that hinder its efficiency. These include the need for multiple components—a reward model, a policy model, a reference policy, and a value model—which significantly increase memory and computational overhead. Additionally, the robustness of RLHF is a concern due to the disjoint training of the reward and policy models. To address these limitations, Direct Preference Optimization (DPO)~\cite{rafailov2024direct} and SLiC~\cite{zhao2022calibrating, zhao2023slic} have emerged as more efficient alternatives. These methods directly fine-tune LLMs using human preference data, bypassing the complexity of RLHF. By optimizing the model through closed-form solutions of preference objectives, DPO and SLiC have shown promise in machine translation tasks, particularly in reducing computational complexity while maintaining strong performance~\cite{zeng2024teaching, wu2024word}.

Despite these advancements, a critical challenge persists in the selection of high-quality preference data, which is essential for offline methods like DPO and SLiC. Recent works such as LLaMA-3~\cite{dubey2024llama} emphasize the importance of careful data selection and cleaning. LLaMA-3 collects preference data from models trained on diverse data mixes, and further refines the data by filtering based on quality, difficulty, and removing semantically redundant sentence pairs. These iterative data processing steps are crucial for preventing distribution shifts and ensuring high-quality data for each round of DPO fine-tuning. However, such exhaustive data cleaning procedures come at the cost of high memory and time complexity, making them less scalable for large-scale translation tasks.

To address these limitations, recent research has explored more flexible and efficient data selection strategies. RSO~\cite{liu2023statistical} proposes a statistical rejection sampling method to subsample preference data from the target optimal policy, effectively estimating the optimal policy distribution. RS-DPO~\cite{khaki2024rs}, on the other hand, utilizes a simpler approach by selecting preference pairs based on the reward difference between sentences. RS-DPO scores a fixed number of responses for each prompt using a point-wise reward model, maintaining only those pairs with reward differences above a predefined threshold. Although these methods improve the efficiency of data selection, they primarily focus on reward values and fail to consider the model's confidence in its predictions, which can be critical for determining which sentence pairs offer the most learning potential.

% Our method, Confidence-Reward driven Preference Optimization (CRPO), aims to address these gaps by jointly considering both rewar scores and model confidence for data selection. As illustrated in Figure~\ref{fig:comp}, CRPO follows a different approach compared to RSO and RS-DPO. While RSO selects candidate translations based on reward scores using statistical rejection sampling, and RS-DPO keeps sentence pairs with large reward differences, CRPO combines reward differences with model confidence. 

Our method, Confidence-Reward driven Preference Optimization (CRPO), as illustrate in Figure~\ref{fig:comp}, aims to address these gaps by jointly considering both reward scores and model confidence for data selection, comparing with RSO that statistically selects candidate translations based on reward scores and RS-DPO that keeps sentence pairs with large reward differences. This approach allows CRPO to select data where the model struggles the most—sentence pairs with high reward differences but where the model is uncertain or incorrect—leading to more effective fine-tuning. By incorporating both of these factors, CRPO ensures that the selected data is not only high-quality but also maximizes the model's learning potential.

While CRPO was primarily designed for LLMs, its application is not limited to decoder-only architectures. Our method also extends to encoder-decoder models, such as NLLB (No Language Left Behind)~\cite{costa2022no}, which have shown strong performance in multilingual translation tasks. NLLB, which is designed to handle over 200 languages, benefits from similar preference optimization techniques, where challenging sentence pairs are selected based on both reward differences and model uncertainty. The success of CRPO on NLLB further demonstrates the method’s versatility, as it effectively addresses the challenges of both LLM-based and encoder-decoder-based translation models.

% \end{itemize}

\section{{Preliminaries}}

% 1.Human Preference Dataset; 2.Reward model; 3.DPO (objective, loss, gradient) 4. CPO

To fine-tune LLMs with human preference annotation on machine translation, translation sentence pair given each source sentence is collected from  reference policy $\pi_{ref}$, larger LLMs such as GPT-4 or human annotator. We define the human preference dataset as $\mathcal{D}_{\succ}=\{x^{(i)}, y^{(i)}_w, y^{(i)}_l\}_{i=1}^N$, where $x^{(i)}$ refers to the $i$th source sentence, $y^{(i)}_w$ and $y^{(i)}_l$ are prefered and disprefered sentence respectively annotated by either human or reward model, and the dataset contains $N$ sentence pairs in total. In this paper, we build a candidate set by sampling $K$ sentences output from reference policy as $\{y^{(ij)}\}_{j=1}^K\sim\pi_{ref}(y|x^{(i)})$ and then score each sentence $y^{(ij)}$ with point-wise reward model as $r^{(ij)}=R(x^{(i)}, y^{(ij)})$. To further construct $\mathcal{D}_{\succ}$, sentence pairs are selected from the candidate set. Two recent preference data selection methods are shown below.

\textbf{RSO}~\cite{liu2023statistical} approximates optimal policy $\pi^*$ with $\pi_{ref}$ using statistical rejection sampling. The expected acceptance rate for each candidate sentence $y^{(ij)}$ is $\mathbb{E}_{y^{(ij)}\sim\pi_{ref}}[\exp(\frac{1}{\beta}\cdot (r^{(ij)}-r_{max}))]$, where $r_{max}$ refers to the maximum reward among candidate sentences and the reward value is the main consideration for acceptance decision.

\textbf{RS-DPO}~\cite{khaki2024rs} calculates the reward difference between each sentence pair and accept a sentence pair when $\sigma(\frac{r^{ij}-r^{il}}{\tau})>\eta$ where $\eta$ is the threshold defined as a hyperparameter and $\sigma(\cdot)$ is the sigmoid function with the formula of $\sigma(x)=\frac{1}{1+e^{-x}}$.

Given the preference dataset, both DPO and RLHF fine-tune LLMs to optimize the following objective:
\begin{equation}
\begin{split}
    \max_{\theta}\mathbb{E}_{x^{(i)}\sim P, y^{(i)}\sim \pi_\theta}[R(x^{(i)},y^{(i)}))]-\\
    \beta\mathbb{D}_{KL}[\pi_\theta(y|x)||\pi_{ref}(y|x)]
    \label{eq:obj}
\end{split}
\end{equation}
where $\theta$ refers to the parameter of current policy $\pi_\theta$ and $\pi_{ref}$ refers to the reference policy. DPO calculates the closed form solution of Equation~\ref{eq:obj} and defines the loss function with Bradley-Terry (BT) model~\cite{bradley1952rank} as:
% preference probability with Bradley-Terry (BT) model~\cite{bradley1952rank} as $\mathbb{P}(y_1\succ y_2|x)=\sigma(r^*(x,y_1)-r^*(x,y_2))$ which indicates the probability that $y_1$ is more preferred than $y_2$ and where $r^*$ refers to the optimal reward value solved from the optimal policy. With reparameterization to $r^*$, DPO fine-tunes $\theta$ by maximizing BT model and defines the loss function to be:
\begin{equation}
    \begin{split}
    &\mathcal{L}_{DPO}(\pi_\theta;\pi_{ref})=-\mathbb{E}_{(x,y_w,y_l)\sim \mathcal{D}_{\succ}}\\
    &[\log\sigma(\beta\log\frac{\pi_\theta(y_w|x)}{\pi_{ref}(y_w|x)}-\beta\log\frac{\pi_\theta(y_l|x)}{\pi_{ref}(y_l|x)})]
    \label{eq:dpo}
    \end{split}
\end{equation}
which directly fine-tunes LLM on the preference dataset.

\citet{xu2024contrastive} propose contrastive preference optimization (CPO) to set the reference policy as uniform prior $U$ for efficiency and act as an upper bound of DPO loss, the format of which is defined to be:
\begin{equation}
\begin{split}
    &\mathcal{L}_{CPO}(\pi_\theta;U)=-\mathbb{E}_{(x,y_w,y_l)\sim \mathcal{D}_{\succ}}\\
    &[\log\sigma(\beta\log\pi_\theta(y_w|x)-\beta\log\pi_\theta(y_l|x)]
    \label{eq:cpo}
\end{split}
\end{equation}
To further enhance LLMs performance, SFT term is added to CPO to clone the behavior of preferred sentences. Moreover, considering the importance of data quality to offline training, \citet{xu2024contrastive} construct preference dataset with translation sentences from GPT4, human annotator and pretrained policy, named as triplet dataset.

Sentences with higher quality or reward could benefit the training of preference optimization, which is the main concern of RSO, RS-DPO and the triplet dataset used in CPO. But the performance of policy $\pi_\theta$ is also important and not considered in these methods. Although RSO jointly leverages the distribution of $\pi_{ref}$ and optimal policy, the acceptance rate is mainly based on reward.

\section{CRPO: Confidence-Reward Driven Preference Optimization}

% DPO -> plus, CPO -> multiply

Instead of reward, we consider the acceptance of sentence pair with the value of optimization loss in two ways, loss value and loss change. A higher loss value indicates that the policy achieves worse performance and the information of related data has not been learnt sufficiently. Similarly, a large loss change during training indicates that the policy extracts useful information from the related data to reduce the prediction confusion or even correct the error prediction. Thus the sentence pairs with either high loss value or loss change are potential to benefit model fine-tuning. In this section, we analyze these two terms on DPO loss and derive two formulations of Confidence-Reward Score (CR-Score) for data selection respectively, Confidence-Reward Plus (CR$+$) to measure loss change and Confidence-Reward Multiplication (CR$\times$) to measure loss value. Although the derivation is different, we will show that both these two scores share the idea of combining model confidence with sentence reward.

\subsection{CR$+$: Derivation from Loss Change}

We start with the derivation from loss change. For the reason that both $\log$ and $\sigma$ are monotonic increasing functions, we simplify the loss change as the difference of minus term inside $\sigma$ function of Equation~\ref{eq:dpo} during training. Formally, taking two parameters $\theta_1$ and $\theta_2$, the loss change is defined as:
\begin{equation}
\begin{split}
    \Delta_\theta\mathcal{L} := &[\log\pi_{\theta_2}(y_w|x)-\log\pi_{\theta_2}(y_l|x)] + \\
    &[\log\pi_{\theta_1}(y_l|x)-\log\pi_{\theta_1}(y_w|x)]
    \label{eq:delta}
\end{split}
\end{equation}
where $\theta_1$ is the parameter before the fine-tuning and we set $\pi_{\theta_1}$ to be $\pi_{ref}$. $\theta_2$ is the parameter after certain steps of fune-tuning, which is hard to be calculated specifically. One potential way to approximate $\pi_{\theta_2}$ is to use the target optimal policy $\pi^*$ and the loss change will be drived into:
\begin{equation}
    \Delta_\theta\mathcal{L} := \frac{R(x, y_w)-R(x, y_l)}{\beta}
\end{equation}
which is exactly the selection metric used in RS-DPO. But in practise, the translation ability is the main concern during inference which is measured by the reward. As also mentioned in \citet{meng2024simpo}, we expect the trained policy to have higher probability to generate high-reward sentence which is different from the distribution of the optimal policy.  Noted that $\Delta_\theta\mathcal{L}$ is defined for data selection which should serve our practical purpose, so we directly approximate $\pi_{\theta_2}$ as the distribution following reward value as:
\begin{equation}
    \pi_{\theta_2}(y|x):=\frac{1}{Z_r(x)}\exp(K\cdot R(x,y))
    \label{eq:rew_ap}
\end{equation}
Compared with the optimal policy, $\pi_{ref}$ is not included and $K$ is a hyperparameter that represents how much we trust the reward model and does not necessarily equal to $\frac{1}{\beta}$. As a result, we define the CR$+$ as $\Delta_\theta\mathcal{L}$ with the formulation derived from Equation~\ref{eq:delta} and Equation~\ref{eq:rew_ap} as:
\begin{equation}
\begin{split}
    \text{CR}+ := &\underbrace{K\cdot [R(x,y_w)-R(x,y_l)]}_{\text{Reward}} + \\
    &\underbrace{[\log\pi_{ref}(y_l|x)-\log\pi_{ref}(y_w|x)]}_{\text{Confidence}}
\end{split}
\end{equation}
where the first term is about reward difference between sentence pair and the second term is the likelihood difference of $\pi_{ref}$ prediction. The larger value of second term indicates that $\pi_{ref}$ has higher confidence on generating $y_l$ rather than $y_w$. Additionally, the two terms in CR$+$ conflict each other that the reward term encourages a larger reward for $y_w$ while the confidence term prefers smaller likelihood of $y_w$ generation. In other word, CR$+$ selects sentence pair with larger reward difference and worse performance for $\pi_{ref}$ to distinguish their qualities. Intuitively, CR$+$ tends to select dispreferred sentences that $\pi_{ref}$ tends to generate but refuses to generate after training and preferred sentences that $\pi_{ref}$ fails to generate but tends to generate after fine-tuning, which thus leads to large behavior change of policy.

\subsection{CR$\times$: Derivation from Loss Value}

We then consider the derivation from loss value. For the reason that in the first iteration $\pi_\theta$ is always set to be $\pi_{ref}$, $\mathcal{L}_{DPO}$ is a constant and cannot be used for data selection. So we base on $\mathcal{L}_{CPO}$ instead to evaluate sentence pairs. Similar to CR$+$, we only consider the minus term inside $\sigma$ function of Equation~\ref{eq:cpo} and construct a more general format as follow:
\begin{equation}
\begin{split}
    \mathcal{L}(\pi_\theta) = &\gamma(R(x,y_w)-R(x, y_l))\cdot \\
    &[\log\pi_\theta(y_w|x)-\log\pi_\theta(y_l|x)]
\end{split}
\end{equation}
where $\gamma(\cdot)$ is a mapping function measuring the correlation between reward score $R(x,y)$ and the translation quality of sentence. For CPO loss, $\gamma(\cdot)$ is set to be the following format:
\begin{equation}
    \gamma(\cdot) = 
        \begin{cases}
         1 & \text{for } R(x,y_w)>R(x,y_l) \\ 
         -1 & \text{otherwise}
        \end{cases}
    \label{eq:sign}
\end{equation}
where the reward model is totally trusted. In this case, a sentence with higher reward is considered to have sufficient quality advantage compared to that with lower reward, regardless of the reward difference. This format of $\gamma(\cdot)$ does not fit our goal for two reasons. Firstly, error exists in reward model and a margin needs to be maintained for reward difference. Secondly, it is not reasonable to force the policy to separate sentence pairs with small reward difference. As a result, we need the $\gamma(\cdot)$ to represent the reward gap in order to measure the quality and trustness of sentence pair. In practise, with point-wise reward model outputting reward within the range of $[0,1]$, we set:
\begin{equation}
    \gamma(R(x,y_w)-R(x,y_l)) = R(x,y_w)-R(x,y_l)
\end{equation}
the value of which also falls within the range of $[0,1]$ when $R(x, y_w)>R(x,y_l)$. Specially, for a ground truth and an irrelevant sentence outputs, the ideal value of $R(x,y_w)-R(x,y_l)$ is closed to 1. For the reason that a larger $\mathcal{L}_{CPO}$ desire a smaller minus term inside $\sigma$ function, we define CR$\times$ as the minus of $\mathcal{L}$ as:
\begin{equation}
\begin{split}
    \text{CR}\times:=&-\mathcal{L}(\pi_{ref})\\
    =&\underbrace{[R(x,y_w)-R(x,y_l)]}_{\text{Reward}}\cdot \\
    &\underbrace{[\log\pi_{ref}(y_l|x)-\log\pi_{ref}(y_w|x)]}_{\text{Confidence}}
\end{split}    
\end{equation}
which is the multiplication of the reward term and the confidence term. Similar to CR$+$, CR$\times$ also tends to select sentence pair with larger reward difference and worse performance of $\pi_{ref}$. Intuitively, CR$\times$ selects sentence pairs with sufficient quality gap while $\pi_{ref}$ fails to distinguish, leading to trustworthy large loss value.

\subsection{Further Discussion}

\textbf{Comparison between CR$+$ and CR$\times$.} Although CR$+$ is derived from loss change and CR$\times$ from loss value, both scores incorporate reward and confidence terms, aiming to maximize the discrepancy between the reward and the policy $\pi_{ref}$. While it is possible to multiply CR$+$ with an additional reward term, as in CR$\times$, this would introduce redundancy, as CR$+$ already contains a reward component. The key distinction between the two lies in the way they handle the magnitude difference between the reward and confidence terms. In CR$+$, this difference necessitates careful tuning of the hyperparameter $K$, which not only adjusts for the reliability of the reward model but also bridges the gap between the reward and confidence scales. In contrast, CR$\times$ naturally balances the two terms through multiplication, eliminating the need for such manual adjustments. In practice, for a specific task and LLM, we estimate $K$ by selecting reward and confidence values that best approximate a balanced contribution from both terms, ensuring robustness across various settings.

\textbf{Why CR-Score?} In machine translation, methods like DPO and CPO have proven effective for fine-tuning LLMs, but the challenge of selecting high-quality preference data remains unresolved. CR-Score offers a systematic approach to evaluate the potential contribution of sentence pairs before the actual model training, thereby guiding more informed data selection. Unlike RS-DPO and RSO, which focus primarily on reward scores, CR-Score incorporates the likelihood of sentence generation by the LLM. This enables the exclusion of "easy" sentence pairs—those where the model already performs well—focusing instead on pairs where the model is uncertain, maximizing the impact of each data point on fine-tuning.

\textbf{CRPO Algorithm.} The algorithm for data selection with CR-Score is outlined in Appendix~\ref{sec:app_alg}. Instead of evaluating all sentence pairs, we begin by selecting the sentence with the highest reward score, $y_w$, to ensure a baseline of sentence quality. This approach, similar to that used in CPO and recent DPO applications (e.g., LLaMA-3~\cite{dubey2024llama}), enhances fine-tuning by prioritizing high-reward sentences. Moreover, we filter out sentence pairs with negative CR-Score, ensuring that only the most informative data is retained in the preference dataset. By combining confidence-reward-driven data selection with preference optimization, CRPO creates a more effective fine-tuning strategy that balances model confidence and reward, enhancing overall model performance.

% \textbf{CRPO Algorithm.} The algorithm for data selection with CR-Score is outlined in Alg~\ref{alg:crpo}. Instead of evaluating all sentence pairs, we begin by selecting the sentence with the highest reward score, $y_w$, to ensure a baseline of sentence quality (line 7). This approach, similar to that used in CPO and recent DPO applications (e.g., LLaMA-3~\cite{dubey2024llama}), enhances fine-tuning by prioritizing high-reward sentences. In lines 9-13, we filter out sentence pairs with negative CR-Score, ensuring that only the most informative data is retained in the preference dataset. By combining confidence-reward-driven data selection with preference optimization, CRPO creates a more effective fine-tuning strategy that balances model confidence and reward, enhancing overall model performance.

\section{Related Works}

% To align LLMs with human preference and enhance their translation ability, instruction tunning with translation instruction has been a popular method to fine-tune LLMs on multilingual datasets~(\cite{yang2023bigtranslate,chen2023improving,zhu2023extrapolating,zhang2023bayling}). But gap still exists when compare LLMs with conventional translation models~(\cite{zhu2023multilingual}). To further improve the performance, RLHF is introduced to fine-tune the language model~(\cite{christiano2017deep}). In order to improve the robustness and efficiency of RLHF, SLiC~(\cite{zhao2022calibrating,zhao2023slic}) and DPO~(\cite{rafailov2024direct}) are proposed to directly train LLMs on preference dataset. SLiC introduces contrastive loss to focus on misclassified sentence pairs and DPO calculates closed-form solution on RLHF object to optimize BT model. CPO~(\cite{xu2024contrastive}) develops upon DPO to release the complexity caused by the requirement of reference model and add additional SFT term to clone the behavoir of preferred sentences. Although these preference optimization methods achieve dramatic success, the offline training strategy causes their sensitivity toward the quality of preference data. To address this problem, we analyze the DPO training process and propose CR-Score to effectively select essential sentence pair to reach the DPO objective.

\textbf{Preference Optimization for Machine Translation.} To align LLMs with human preference and enhance their translation ability, RLHF is introduced to fine-tune the language model~\cite{christiano2017deep}. In order to improve the robustness and efficiency of RLHF, DPO~\cite{rafailov2024direct} calculates closed-form solution on RLHF object to optimize BT model and directly train LLMs on preference dataset. CPO~\cite{xu2024contrastive} develops upon DPO to release the complexity caused by the requirement of reference model and add SFT term for behavior cloning. Although these preference optimization methods achieve dramatic success, the offline training strategy causes their sensitivity toward the quality of preference data. To address this problem, we analyze the loss value and loss change and propose CR-Score to effectively select essential sentence pair to reach the DPO objective.

\textbf{Rejection Sampling.} To select preference data for alignment, rejection sampling is a widely adopted method. RSO~\cite{liu2023statistical} introduces statistical rejection sampling~\cite{neal2003slice} and sample preference sentences from target policy distrbution. ReST~\cite{gulcehre2023reinforced} iteratively increase reward threshold and apply rejection sampling to select higher quality sentence for further RLHF step. RS-DPO~\cite{khaki2024rs} instead sets the threshold of reward difference between sentence pairs and only maintains those with large enough preference difference. Specifically for the machine translation task, the MBR score~\cite{yang2023direct, finkelstein2023mbr} leverages reference-based metric to estimate the expected utility of each candidate translation in relation to the set of pseudo-references. To reduce the computational complexity, \citet{finkelstein2023mbr} further consider to score translations with QE metric and fine-tune LLM with the best translation result. However, these sampling methods only focus on reward value neglecting the performance of the pretrained policy. Instead, our proposed CR-Score considers reward and policy confidence together.

\section{Experiments}
\label{sec:exp}

We evaluate CRPO on machine translation task and compare it with five baselines, evaluated with COMET and BLEURT~\cite{sellam2020bleurt} metrics. Moreover, we adopt ablation studies to consider more data selection strategies and the effect of reward and confidence term on CR-Score.

\subsection{Dataset}
Following CPO~\cite{xu2024contrastive}, we consider 10 translation directions in this paper: $en\leftrightarrow zh$, $en\leftrightarrow de$, $en\leftrightarrow cs$, $en\leftrightarrow is$, $en\leftrightarrow ru$. Our preference training dataset of machine translation task is derived from FLORES-200 dataset~\cite{costa2022no} with the same source sentences applied to fine-tune ALMA~\cite{xu2023paradigm} in CPO. In the training dataset, 3,065 source sentences are contained in each of $en\leftrightarrow zh$ and $en\leftrightarrow de$, and 2,009 source sentences are contained in each of other translation directions. In total, 24,314 source sentences are included. Note that ALMA is also pretrained on a subset of FLORES-200 dataset, we collect 64 candidate sentences for each source sentence with the pretrained ALMA to release distribution shift problem which results in 784,640 candidate translation sentences. The sampling temperature is set to be 0.9 and top-p is set to be 0.9. To evaluate the quality of translation sentences during preference dataset construction, we use two 3.5B COMET models, \textit{Unbabel/XCOMET-XL}~\cite{guerreiro2023xcomet} and \textit{Unbabel/wmt23-cometkiwi-da-xl}~\cite{rei2023scaling}, as reward models and average the two output scores from them as the final reward of translation sentences. Moreover, we follow CPO to extract data of $en\leftrightarrow is$ from WMT21~\cite{freitag2021results} and data of the other 8 translation directions from WMT22~\cite{freitag2022results} as test set, resulting in 17,471 translation pairs in total. 

\subsection{Experiment Setup}
We train the ALMA-7B in a many-to-many multilingual translation manner, starting with ALMA-7B-Pretrain-LoRA as initial checkpoint. Then we sample preference dataset with CR-Score and apply DPO to fine-tune the pretrained ALMA-7B model on preference dataset. Then we evaluate the translation results with COMET models. Besides the reward models XCOMET and KIWI-XL, we also utilize COMET-22 (\textit{Unbabel/wmt22-comet-da}) and KIWI-22 (\textit{Unbabel/wmt22-cometkiwi-da}) for fair comparison which are not involved in either data selection or model fine-tuning. 
% For simplicity, we abbreviate KIWI-22, COMET-22, XCOMET and KIWI-XL as K-22, C-22, XC and K-XL respectively. 
During inference, we generate the final output of ALMA-7B with beam search, setting beam size to be 5 and maximum sequence length to be 512 tokens. For more details, please refer to the Appendix~\ref{sec:exp_detail}.

% \begin{table*}[t]
%     % \setlength\tabcolsep{1.5pt}
%     \centering
%     \caption{Average results on ten translation directions. \textbf{Black bold font} refers to the best result and \textbf{gray bold font} refers to the second best result.}
%     \label{tab:avg}
%     \begin{tabular}{l|cccc}
%     \hline
%         \multirow{2}{*}{\textbf{Method}} & \multicolumn{4}{c}{Average} \\\cline{2-5}
%          & \textbf{KIWI-22} & \textbf{COMET-22} & \textbf{XCOMET} & \textbf{KIWI-XL} \\\hline
%         SFT & 0.8149 & 0.8563 & 0.9243 & 0.7338 \\\hline
%         RSO & 0.8197 & 0.8598 & 0.9277 & 0.7403 \\
%         RS-DPO ($\eta$=0.60/0.50) & 0.8134 & 0.8547 & 0.9189 & 0.7299 \\
%         RS-DPO ($\eta$=0.65/0.55) & 0.8140 & 0.8553 & 0.9205 & 0.7311 \\
%         Triplet Dataset & 0.8168 & 0.8581 & 0.9274 & 0.7371 \\\hline
%         CRPO$+$ & \textbf{0.8218} & \textbf{0.8618} & \textbf{0.9311} & \textbf{0.7462} \\
%         CRPO$\times$ & \textcolor{gray}{\textbf{0.8217}} & \textcolor{gray}{\textbf{0.8612}} & \textcolor{gray}{\textbf{0.9307}} & \textcolor{gray}{\textbf{0.7451}} \\\hline
%     \end{tabular}
% \end{table*}

\subsection{Baselines}

We compare CRPO with five baselines, QE Fine-tuning~\cite{finkelstein2023mbr}, RSO~\cite{liu2023statistical}, RS-DPO~\cite{khaki2024rs}, MBR Score~\cite{yang2023direct} and Triplet dataset~\cite{xu2024contrastive}. As an additional comparison, we also calculate the evaluation score of gold reference sentences from the WMT dataset.

% \textbf{SFT}. We fine-tune policy with the gold reference sentences.

\textbf{QE Fine-tuning}. We choose the sentence with the highest QE reward score from the candidate set as the target sentence to fine-tune policy.

\textbf{RSO}. We statistically sub-sample 8 sentences from the candidate dataset. The acceptance rate for each sentence is $\exp(\frac{1}{\beta}\cdot (r^{(ij)}-r_{max}))$. 

% In order to construct preference dataset, we calculate the sentence with largest reward as preferred sentence and the sentence with smallest reward as dispreferred sentence.

\textbf{RS-DPO}. For convenience in hyperparameter tuning, we replace $\sigma(\frac{r_1-r_2}{\tau})$ with $r_1-r_2$ for RS-DPO. For the reason that the translation direction task $* \rightarrow en$ is harder than $en\rightarrow *$, we set a larger $\eta$ for the former cases. In general, we consider two groups of values for $\eta$ for a fair comparison, specifically 0.6 for $en \rightarrow *$ and 0.5 for $* \rightarrow en$, 0.65 for $en \rightarrow *$ and 0.55 for $* \rightarrow en$.

\textbf{MBR Score}. We calculate MBR score for candidate translation sentences with BLEURT-20~\cite{sellam2020bleurt} Metric. Specifically, we consider \textbf{MBR-BW} as selecting the best and worst translation sentences and \textbf{MBR-BMW} as selecting the best, middle, and worst translation sentences.

\textbf{Triplet Dataset}. We reuse the preference dataset from Triplet Dataset that is used for CPO training. 

% For each source sentence, Triplet Dataset contains three translation sentences from pretrained ALMA-7B, GPT4 and reference result. We pick the sentence with highest reward model score as preferred sentence and that with smallest reward model score as dispreferred sentence.

\begin{figure*}[t!]
    \centering
    \begin{subfigure}{\textwidth}
        \centering
        \includegraphics[width=\textwidth]{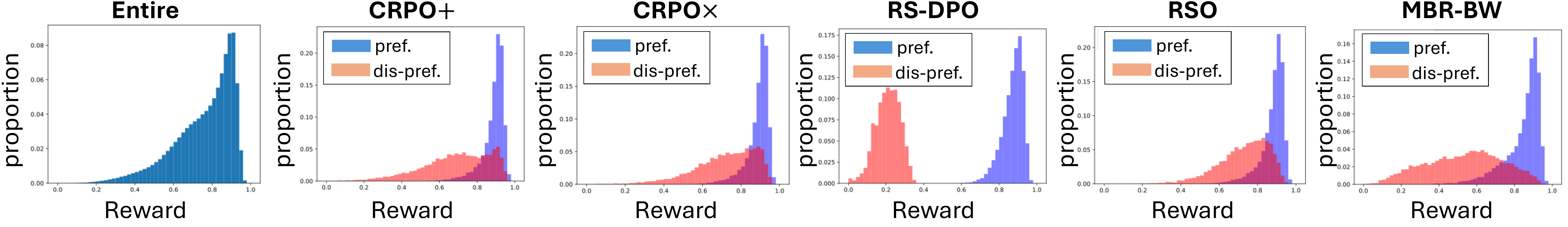}
        \caption{Reward Score.}
        \vspace{0.1cm}
    \end{subfigure}
    \begin{subfigure}{\textwidth}
        \centering
        \includegraphics[width=\textwidth]{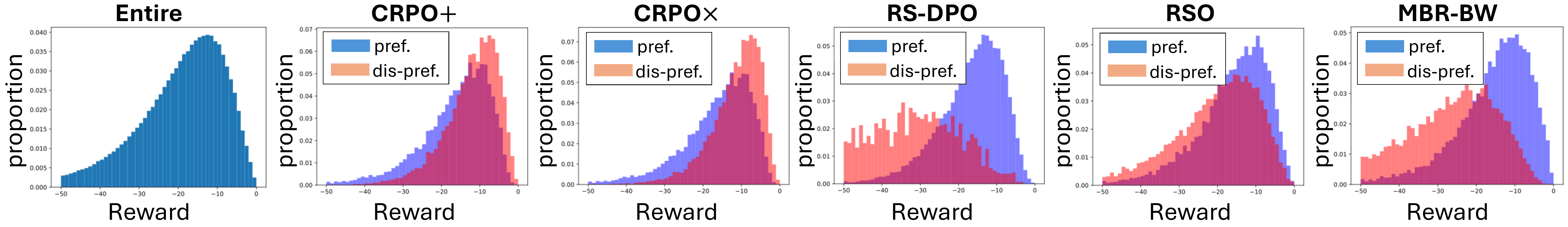}
        \caption{$\log\pi_{ref}$.}
    \end{subfigure}
    \caption{Visualization of reward score and $\log\pi_{ref}$ for the entire candidate dataset, as well as preferred and dis-preferred sentences selected by CRPO$+$, CRPO$\times$, RS-DPO, RSO and MBR-BW.}
    \label{fig:vis_dist}
\end{figure*}

\subsection{Experiment Results}

\begin{table}[t]
    \setlength\tabcolsep{2.pt}
    \small
    \centering
    \caption{Average results on ten translation directions. \textbf{Black bold font} refers to the best result and \textcolor{gray}{\textbf{gray bold font}} refers to the second best result. RS-DPO-1 (RS-DPO-2) refers to RS-DPO with $\eta=0.6$ ($\eta=0.65$) for $en\rightarrow *$ and $\eta=0.5$ ($\eta=0.55$) for $*\rightarrow en$.}
    \label{tab:avg}
    \begin{tabular}{l|cccc}
    \hline
        % \multirow{2}{*}{\textbf{Method}} & \multicolumn{4}{c}{Average} \\\cline{2-5}
        %  & \textbf{K-22} & \textbf{C-22} & \textbf{XC} & \textbf{K-XL} \\\hline
        \multirow{2}{*}{\textbf{Method}} & \multicolumn{4}{c}{\textbf{Average}} \\\cline{2-5}
         & \textbf{KIWI22} & \textbf{COMET22} & \textbf{XCOMET} & \textbf{KIWI-XL} \\\hline
        ALMA-7B & 0.8140 & 0.8559 & 0.9203 & 0.7306 \\
        % SFT &  &  &  &  \\
        QE Ft. & 0.8149 & 0.8563 & 0.9243 & 0.7338 \\
        Goal Ref. & 0.8098 & - & 0.9118 & 0.7268 \\\hline 
        % GPT-4o & \textbf{0.8243} & \textbf{0.8725} & \textcolor{gray}{\textbf{0.9294}} & \textbf{0.7476} \\ \hline
        RSO & 0.8197 & 0.8598 & 0.9277 & 0.7403 \\
        RS-DPO-1 & 0.8134 & 0.8547 & 0.9189 & 0.7299 \\
        RS-DPO-2 & 0.8140 & 0.8553 & 0.9205 & 0.7311 \\
        Triplet & 0.8168 & 0.8581 & 0.9274 & 0.7371 \\
        MBR-BW & 0.8174 & 0.8589 & 0.9240 & 0.7357 \\
        MBR-BMW & 0.8167 &  0.8588 & 0.9248 & 0.7356  \\\hline
        CRPO$+$ & \textcolor{black}{\textbf{0.8218}} & \textcolor{black}{\textbf{0.8618}} & \textbf{0.9311} & \textcolor{black}{\textbf{0.7462}} \\
        CRPO$\times$ & \textcolor{gray}{\textbf{0.8217}} & \textcolor{gray}{\textbf{0.8612}} & \textcolor{gray}{\textbf{0.9307}} & \textcolor{gray}{\textbf{0.7451}} \\\hline
    \end{tabular}
\end{table}

The average results for ten translation directions are shown in Table~\ref{tab:avg}, where CRPO with CR$+$ achieves the best performance and CR$\times$ gets the second best results. As RSO, RS-DPO and MBR Score mainly select the preference dataset based on sentence reward, the evaluation results emphasize the benefit of adding the confidence term to consider policy behavior. Triplet dataset performs worse than RSO and CRPO, which is mainly caused by distribution shift between $\pi_{ref}$ and response sentences from other resources. Although RS-DPO also constructs preference dataset from candidate sentences, the main reason for its worse performance we think is the performance gap of policy among different translation directions even when we already set different $\eta$ for $en\rightarrow *$ and $*\rightarrow en$. For example, setting $\eta=0.6$, on average around three sentence pairs will be maintained for each source sentence in $en\rightarrow zh$ direction while only $20\%$ of source sentences are maintained in $en\rightarrow de$ direction. A large $\eta$ causes the lack of information in difficult translation directions while small $\eta$ maintains relatively useless information in easy translation directions. A potential solution is to set specific $\eta$ for each translation direction while leading to higher computational cost. On comparison, although hyperparameter $K$ also need to be set in CR$+$, we only need to consider the magnitude gap between reward and confidence and set one value for all translation directions which is more straightforward. We show more results in Appendix~\ref{sec:appendix_exp}, where CRPO achieves the best performance on almost all translation directions, which empirically proves the robustness of our method and the significant role of confidence term. Additionally, we evaluate CRPO with non-COMET family metric BLEURT in Appendix~\ref{sec:bleurt} to address the concern of correlation between reward model and evaluation metric and show that CRPO also achieves the best result.

For further comparison, we visualize the distribution of reward score and $\log\pi_{ref}$ for preferred and dis-preferred sentences selected by different methods in $en\rightarrow *$ translation directions in Figure~\ref{fig:vis_dist} where preferred sentences always have better reward scores. Specifically for RS-DPO, as only sentence pairs with high reward gap are selected, the reward difference of sentence pairs are more obvious than other methods. However, preferred sentences tend to have higher generation likelihood than dis-preferred sentences, leading to "easy" data that model already performs well. Similar problem also exists in RSO and MBR-BW. In comparison, CRPO$+$ and CRPO$\times$ select preferred sentences with worse $\log\pi_{ref}$ and dispreferred sentences with higher $\log\pi_{ref}$ which are more difficult for policy and have better impact to model fine-tuning.

\subsection{Experiment for NLLB}
To evaluate the generalization of CRPO, we extend CRPO to encoder-decoder model, NLLB-1.3B~\cite{costa2022no}, and compare CRPO with RSO, RS-DPO ($\eta=0.82$), Triplet dataset and MBR score. Similar to the setting of ALMA experiment, for the 10 translation directions, we leverage the pretrained NLLB model (checkpoint \textit{facebook/nllb-200-1.3B}) to collect 64 candidate sentences for each source sentence from FLORES-200 dataset and then apply data selection methods to construct preference dataset. We fine-tune NLLB model with DPO and evaluate it on the same WMT dataset. For more details of experiment setting, please refer to the Appendix. 

\begin{table}[t]
    \setlength\tabcolsep{2.pt}
    \small
    \centering
    \caption{Average results on ten translation directions for NLLB. $\eta$ is set to be 0.82 for RS-DPO.}
    \label{tab:avg_nllb}
    \begin{tabular}{l|cccc}
    \hline
        % \multirow{2}{*}{\textbf{Method}} & \multicolumn{4}{c}{Average} \\\cline{2-5}
        %  & \textbf{K-22} & \textbf{C-22} & \textbf{XC} & \textbf{K-XL} \\\hline
        \multirow{2}{*}{\textbf{Method}} & \multicolumn{4}{c}{\textbf{Average}} \\\cline{2-5}
         & \textbf{KIWI22} & \textbf{COMET22} & \textbf{XCOMET} & \textbf{KIWI-XL} \\\hline
        NLLB-1.3B & 0.8009 & 0.8362 & 0.8947 & 0.7001  \\
        QE Ft. & 0.7890 & 0.8201 & 0.8670 & 0.6770 \\
        % QE Ft &  & &  &  \\
        Goal Ref. & 0.8098 & - & \textbf{0.9118} & \textbf{0.7268} \\\hline
        % GPT-4o & 0.8243 & 0.8725 & 0.9294 & 0.7476 \\\hline
        RSO & 0.8142 & 0.8466 & 0.9066 & 0.7183 \\
        RS-DPO & 0.8078 & 0.8406 & 0.8972 & 0.7084 \\
        Triplet & 0.8138 & 0.8465 & 0.9025 & 0.7163 \\
        MBR-BW & 0.8104 & 0.8447 & 0.9026 & 0.7134\\
        MBR-BMW & 0.8073 & 0.8421 & 0.9005 & 0.7080 \\\hline
        CRPO$+$ & \textbf{0.8149} & \textbf{0.8469} & \textcolor{gray}{\textbf{0.9074}} & \textcolor{gray}{\textbf{0.7207}} \\
        CRPO$\times$ & \textcolor{gray}{\textbf{0.8148}} & \textcolor{gray}{\textbf{0.8467}} & 0.9063 & 0.7202 \\\hline
    \end{tabular}
\end{table}

The experimental results are shown in Table~\ref{tab:avg_nllb}. Although worse than goal reference translation results for XCOMET and KIWI-XL scores, CRPO$+$ achieves better performance than other data selection methods, which indicates that CRPO generalizes well to encoder-decoder translation model.

\subsection{Ablation Study}

% \begin{table*}[t]
%     \centering
%     % \setlength\tabcolsep{1.5pt}
%     \caption{Average results on ten translation directions for ablation study.}
%     \label{tab:abl}
%     \begin{tabular}{l|cccc}
%     \hline
%         \multirow{2}{*}{\textbf{Method}} & \multicolumn{4}{c}{Average} \\\cline{2-5}
%          & \textbf{KIWI-22} & \textbf{COMET-22} & \textbf{XCOMET} & \textbf{KIWI-XL} \\\hline
%         SFT & 0.8149 & 0.8563 & 0.9243 & 0.7338 \\\hline
%         MinMaxReward & 0.8194 & 0.8594 & 0.9251 & 0.7387 \\
%         MinMaxProb & 0.8178 & 0.8592 & 0.9265 & 0.7371 \\
%         MinMaxProbOnly & 0.8081 & 0.8508 & 0.9137 & 0.7184 \\
%         TopScores & 0.8183 & 0.8592 & 0.9260 & 0.7380 \\\hline
%         CRPO$+$ & \textbf{0.8218} & \textbf{0.8618} & \textbf{0.9311} & \textbf{0.7462} \\
%         CRPO$\times$ & \textcolor{gray}{\textbf{0.8217}} & \textcolor{gray}{\textbf{0.8612}} & \textcolor{gray}{\textbf{0.9307}} & \textcolor{gray}{\textbf{0.7451}} \\\hline
%     \end{tabular}
% \end{table*}

\begin{table}[t]
    \centering
    \small
    \setlength\tabcolsep{2.pt}
    \caption{Average results on ten translation directions for ablation study.}
    \label{tab:abl}
    \begin{tabular}{l|cccc}
    \hline
        % \multirow{2}{*}{\textbf{Method}} & \multicolumn{4}{c}{Average} \\\cline{2-5}
        %  & \textbf{K-22} & \textbf{C-22} & \textbf{XC} & \textbf{K-XL} \\\hline
        \multirow{2}{*}{\textbf{Method}} & \multicolumn{4}{c}{\textbf{Average}} \\\cline{2-5}
         & \textbf{KIWI22} & \textbf{COMET22} & \textbf{XCOMET} & \textbf{KIWI-XL} \\\hline
        ALMA-7B & 0.8140 & 0.8559 & 0.9203 & 0.7306 \\
        QE Ft. & 0.8149 & 0.8563 & 0.9243 & 0.7338 \\\hline
        MinMaxR & 0.8194 & 0.8594 & 0.9251 & 0.7387 \\
        MinMaxP & 0.8178 & 0.8592 & 0.9265 & 0.7371 \\
        MinMaxPO & 0.8081 & 0.8508 & 0.9137 & 0.7184 \\
        TopScores & 0.8183 & 0.8592 & 0.9260 & 0.7380 \\\hline
        CRPO$+$ & \textbf{0.8218} & \textbf{0.8618} & \textbf{0.9311} & \textbf{0.7462} \\
        CRPO$\times$ & \textcolor{gray}{\textbf{0.8217}} & \textcolor{gray}{\textbf{0.8612}} & \textcolor{gray}{\textbf{0.9307}} & \textcolor{gray}{\textbf{0.7451}} \\\hline
    \end{tabular}
\end{table}

% \begin{table}[t]
%     \centering
%     \setlength\tabcolsep{2.5pt}
%     \caption{Average results of CRPO on mixed dataset.}
%     \label{tab:mix}
%     \begin{tabular}{l|cccc}
%     \hline
%         \multirow{2}{*}{\textbf{Method}} & \multicolumn{4}{c}{Average} \\\cline{2-5}
%          & \textbf{K-22} & \textbf{C-22} & \textbf{XC} & \textbf{K-XL} \\\hline
%         CRPO$+$ & 0.8218 & 0.8618 & \textcolor{gray}{\textbf{0.9311}} & \textcolor{gray}{\textbf{0.7462}} \\
%         CRPO$\times$ & 0.8217 & 0.8612 & 0.9307 & 0.7451 \\\hline
%         Triplet & 0.8168 & 0.8581 & 0.9274 & 0.7371 \\
%         CRPO$+$(mixed) & \textbf{0.8223} & \textcolor{gray}{\textbf{0.8622}} & 0.9299 & 0.7458 \\
%         CRPO$\times$(mixed) & \textcolor{gray}{\textbf{0.8221}} & \textbf{0.8626} & \textbf{0.9319} & \textbf{0.7465} \\\hline
%     \end{tabular}
% \end{table}

% \begin{table}[t]
%     \centering
%     \setlength\tabcolsep{2.5pt}
%     \caption{Average results of CRPO and Triplet Dataset on CPO.}
%     \label{tab:cpo}
%     \begin{tabular}{l|cccc}
%     \hline
%         \multirow{2}{*}{\textbf{Method}} & \multicolumn{4}{c}{Average} \\\cline{2-5}
%          & \textbf{K-22} & \textbf{C-22} & \textbf{XC} & \textbf{K-XL} \\\hline
%         Triplet (CPO) & 0.8175 & 0.8587 & 0.9284 & 0.7389 \\
%         CRPO$+$ (CPO) & \textbf{0.8214} & \textbf{0.8607} & \textbf{0.9306} & \textbf{0.7451} \\
%         CRPO$\times$ (CPO) & \textbf{0.8214} & 0.8604 & 0.9302 & 0.7450 \\\hline
%     \end{tabular}
% \end{table}

In the ablation study, we compare CRPO with more data selection methods to evaluate the effect of reward term and confidence term. We consider some questions that can be potentially raised from CRPO in the Appendix~\ref{sec:app_qa}. Specifically, we consider the following methods:

\textbf{MinMaxR.} To further evaluate the contribution of confidence term in CR-Score, we drop it from the CR$+$ and only maintain the reward difference as the score. In another word, the sentence with maximum reward score is selected as preferred sentence and the sentence with minimum reward score is selected as dispreferred sentence.

\textbf{MinMaxP.} Similarly, we evaluate the contribution of reward term by setting $K=0$ in CR$+$ and selecting the sentence pair with maximum value of $[\log\pi_{ref}(y_l|x)-\log\pi_{ref}(y_w|x)]$. For fair comparison, the sentence pairs resulting in negative value are dropped. Noted that sentence pair with higher value of MinMaxP gets larger CPO loss value.

\textbf{MinMaxPO.} As a comparison with MinMaxP, we select the sentence with maximum likelihood and the sentence with minimum likelihood, among which the sentence with higher reward is set as $y_w$.

\textbf{TopScores.} RSO utilizes statistical reject sampling to sample sentences from candidate set. We instead consider reject sampling directly based on the acceptance rate, that is only keeping sentences with top reward scores. Among these selected sentences, we set the one with highest reward score as preferred sentence and the one with smallest reward score as dispreferred sentence.

The average experiment results on ten translation directions are shown in Table~\ref{tab:abl}. Dropping confidence term from CRPO, MinMaxR only considers the sentence quality from reward while neglecting the information policy could learn from related sentences, resulting in worse performance compared with CRPO. On the contrary, MinMaxP and MinMaxPO ensure the diversity of sentences by likelihood difference while neglecting the sentence quality, also gets lower evaluation scores. The above experiments empirically prove the necessity of combining sentence reward with prediction likelihood for sentence selection. Comparing with RSO, TopScores selects sentence with top scores rather than applying statistical rejection sampling and gets worse results, the reason of which might related to sentence diversity we think.

\section{Conclusion}

In this paper, we argue that sentence pairs with large loss value or loss change during training contains information the model has not yet learn and could benefit the model fine-tuning. We analyze loss value and loss change based on DPO and find them to be controlled by confidence terms measuring the prediction likelihood difference and reward terms measuring reward difference of sentence pair, based on which CR$+$ and CR$\times$ are designed for data selection. With experiment results, we empirically prove that CRPO outperforms previous data sampling method in machine translation tasks. And based on ablation study, we show the necessity of considering the two terms together for fine-tuning.

\section*{Limitation}
In CRPO, we only select the sentence pair with maximum CR-Score, which will discard high quality data with slightly smaller CR-Score. Potential solution to this limitation includes leveraging preset threshold, or CR-Score distribution analysis.

\bibliography{custom}

\appendix

\clearpage

\newpage

\section{Algorithm}
\label{sec:app_alg}

The algorithm for data selection with CR-Score is outlined in Alg~\ref{alg:crpo} where the inputs are source sentence, $\pi_{ref}$ and reward model. In line 3-6, we sample candidate translation sentences from $\pi_{ref}$ and then calculate their reward score and generation likelihood. In line 7, we select the sentence with highest reward score as preferred sentence $y_w$, to ensure a baseline of sentence quality. In lines 9-13, we filter out sentence pairs with negative CR-Score and select the sentence pair with maximum CR-Score, ensuring that only the most informative data is retained in the preference dataset.

\begin{algorithm}[h]
    \caption{Data Selection with Confidence-Reward Score}
    \label{alg:crpo}
    \begin{algorithmic}[1]
        \STATE {\bfseries Input:} Source Sentence $x^{(i)}$, Reference Policy $\pi_{ref}$, Reward Model $R$.
        \STATE {\bfseries Output:} Preference Data $\mathcal{D}_{\succ}^{(i)}$.
        \STATE Set $\mathcal{D}_{\succ}^{(i)}$ as empty, Set $s_{best}=0$;
        \STATE Sample candidate sentences: \\ $\mathcal{Y}=\{y^{(ij)}\}_{j=1}^K\sim\pi_{ref}(y|x^{(i)})$;
        \STATE Collect probabilities: \\ $\mathcal{P}=\{p^{(ij)}=\pi_{ref}(y^{(ij)|x})\}_{j=1}^K$;
        \STATE Collect rewards: \\ $\mathcal{R}=\{r^{(ij)}=R(x^{(i)}, y^{(ij)})\}_{j=1}^K$;
        \STATE $j_{max}=\arg\max_{j} \mathcal{R}$;
        \FOR{each sentence $y^{(ij)}$ in $\mathcal{Y}$}
        \IF{$p^{(ij)}-p^{(ij_{max})}+\epsilon>0$}
        \STATE Calculate CR-Score for ($y^{ij_{max}}, y^{ij}$): \\$s=\text{CR}+$ or $s=\text{CR}\times$;
        \IF{$s>s_{best}$}
        \STATE $\mathcal{D}_{\succ}^{(i)}=(x^{(i)},y^{(ij_{max})},y^{(ij)})$;
        \ENDIF
        \ENDIF
        \ENDFOR
        \STATE Return $\mathcal{D}_{\succ}^{(i)}$;
    \end{algorithmic}
\end{algorithm}

\section{Experimental Details}
\label{sec:exp_detail}

In this section, we provide more details for experiment setting of ALMA and NLLB fine-tuning.

\subsection{Experiment Setting for ALMA}

During the training phase, we train the ALMA-7B in a many-to-many multilingual translation manner, starting with ALMA-7B-Pretrain-LoRA as initial checkpoint. For model fine-tuning, we focus on updating the weights of added LoRA parameters which have a rank of 16 and only add an additional 7.7M parameters to the original 7B size of the model. We follow \cite{xu2024contrastive} to set the default $\beta$ value to be 0.1, the batch size of ALMA-7B in fine-tuning process to be 16, a warm-up ratio to be 0.01 and learning rate to be 0.0001. For all experiments, ALMA-7B is fine-tuned with one single epoch and the maximum length of accommodating sequence is set to be 512 tokens. For CR$+$, we set $K$ to be 50 to bridge the magnitude different between reward and confidence terms. To optimize training efficiency, we implement the model fine-tuning with deepspeed tool~\cite{rasley2020deepspeed}. For machine translation, we follow ALMA to set the prompt as \textit{``Translate this from $\textlangle$source language$\textrangle$ to $\textlangle$target language$\textrangle$; $\textlangle$source language$\textrangle$: $\textlangle$source sentence$\textrangle$; $\textlangle$target language$\textrangle$:''} and exclude the token sequence of prompt for loss calculation. Moreover, during training, we follow \cite{dubey2024llama} to drop the EOS tokens at the end of translation outputs for both preferred and dispreferred sentences to avoid repeated tail content and add SFT term for DPO with coefficient weight set to be 1 for performance enhancement.

\subsection{Experiment Setting for NLLB}

We also train NLLB-1.3B in many-to-many multilingual translation manner, starting with facebook/nllb-200-1.3B as initial checkpoint. During fine-tuning, we add LoRA parameters to liner layers of NLLB model with additional 27.7M parameters which are the only trainable parameters. Similar to ALMA, we set the $\beta$ value for DPO as 0.1, the batch size in fine-tuning process to be 16, a warm-up ratio to be 0.01 and learning rate to be 0.0001. For CR$+$, we again set $K$ to be 50 to bridge the magnitude difference between reward and confidence terms. Following \cite{costa2022no}, we manually set the BOS tokens for both encoder input sentence and decoder output sentence as related language index. We also add SFT term for DPO with coefficient weight set to be 1 for fair comparison. But different from ALMA, we retain the EOS tokens for preference sentence pairs which would not cause the tail content redundancy problem as in ALMA.

\section{More Experimental Results}
\label{sec:appendix_exp}

\begin{table*}[t]
    \centering
    \small
    \setlength\tabcolsep{2.5pt}
    \caption{Experiment results on translation directions of $en\rightarrow *$. The average result over these 5 translation directions are shown in the column of $en\rightarrow *$.}
    \label{tab:ento}
    \begin{tabular}{l|cccc|cccc}
        \hline
        \multirow{2}{*}{\textbf{Method}} & \multicolumn{4}{c|}{$en\rightarrow zh$} & \multicolumn{4}{c}{$en\rightarrow de$} \\\cline{2-9}
        % & \textbf{K-22} & \textbf{C-22} & \textbf{XC} & \textbf{K-XL} & \textbf{K-22} & \textbf{C-22} & \textbf{XC} & \textbf{K-XL} \\\hline
        & \textbf{KIWI22} & \textbf{COMET22} & \textbf{XCOMET} & \textbf{KIWI-XL} & \textbf{KIWI22} & \textbf{COMET22} & \textbf{XCOMET} & \textbf{KIWI-XL} \\\hline
        QE Ft. & 0.8099 & 0.8596 & 0.8611 & 0.7387 & 0.8265 & 0.8574 & 0.9645 & 0.7392  \\
        Goal Ref. & 0.8093 & - & 0.8594 & 0.7402 & 0.8268 & - & 0.9640 & 0.7475 \\ \hline
        RSO & 0.8160 & 0.8631 & 0.8694 & 0.7488 & 0.8310 & 0.8604 & 0.9663 & 0.7456 \\
        RS-DPO ($\eta$=0.60/0.50) & 0.8081 & 0.8574 & 0.8556 & 0.7339 & 0.8265 & 0.8558 & 0.9631 & 0.7369 \\
        RS-DPO ($\eta$=0.65/0.55) & 0.8069 & 0.8564 & 0.8560 & 0.7314 & 0.8254 & 0.8546 & 0.9626 & 0.7345 \\
        Triplet Dataset & 0.8089 & 0.8581 & 0.8690 & 0.7373 & 0.8278 & 0.8591 & 0.9663 & 0.7424 \\
        MBR-BW & 0.8106 & 0.8596 & 0.8571 & 0.7371 & 0.8287 & 0.8588 & 0.9636 & 0.7422 \\
        MBR-BMW & 0.8098 & 0.8604 & 0.8613 & 0.7392 & 0.8280 & 0.8595 & 0.9662 & 0.7419 \\\hline
        CRPO$+$ & \textbf{0.8194} & \textbf{0.8656} & \textcolor{gray}{\textbf{0.8706}} & \textcolor{gray}{\textbf{0.7548}} & \textcolor{gray}{\textbf{0.8326}} & \textcolor{gray}{\textbf{0.8622}} & \textcolor{gray}{\textbf{0.9673}} &\textbf{0.7533} \\
        CRPO$\times$ & \textcolor{gray}{\textbf{0.8178}} & \textcolor{gray}{\textbf{0.8639}} & \textbf{0.8719} & \textbf{0.7555} & \textbf{0.8329} & \textbf{0.8637} & \textbf{0.9675} &\textcolor{gray}{\textbf{0.7530}} \\\hline
    \end{tabular}

    \begin{tabular}{cc}
    \end{tabular}
    
    \begin{tabular}{l|cccc|cccc}
    \hline
        \multirow{2}{*}{\textbf{Method}} & \multicolumn{4}{c|}{$en\rightarrow cs$} & \multicolumn{4}{c}{$en\rightarrow is$} \\\cline{2-9}
        % & \textbf{K-22} & \textbf{C-22} & \textbf{XC} & \textbf{K-XL} & \textbf{K-22} & \textbf{C-22} & \textbf{XC} & \textbf{K-XL} \\\hline
        & \textbf{KIWI22} & \textbf{COMET22} & \textbf{XCOMET} & \textbf{KIWI-XL} & \textbf{KIWI22} & \textbf{COMET22} & \textbf{XCOMET} & \textbf{KIWI-XL} \\\hline
        QE Ft. & 0.8397 & 0.8941 & 0.9246 & 0.7556 & 0.8124 & 0.8574 & 0.9000 & 0.7557  \\
        Goal Ref. & 0.8319 & - & 0.8964 & 0.7405 & 0.8051 & - & 0.8872 & 0.7559 \\ \hline
        RSO & 0.8455 & 0.8980 & 0.9292 & 0.7648 & 0.8135 & 0.8584 & 0.9008 & 0.7602 \\
        RS-DPO ($\eta$=0.60/0.50) & 0.8369 & 0.8923 & 0.9151 & 0.7491 & 0.8108 & 0.8527 & 0.8890 & 0.7499 \\
        RS-DPO ($\eta$=0.65/0.55) & 0.8389 & 0.8931 & 0.9194 & 0.7522 & 0.8100 & 0.8549 & 0.8923 & 0.7533 \\
        Triplet Dataset & 0.8421 & 0.8957 & 0.9306 & 0.7620 & 0.8121 & 0.8575 & 0.8968 & 0.7562 \\
        MBR-BW & 0.8431 & 0.8983 & 0.9232 & 0.7585 & 0.8161 & 0.8596 & 0.8994 & 0.7599 \\
        MBR-BMW & 0.8408 & 0.8964 & 0.9256 & 0.7568 & 0.8161 & 0.8601 & 0.9020 & 0.7608 \\\hline
        CRPO$+$ & \textcolor{gray}{\textbf{0.8483}} & \textbf{0.9019} & \textbf{0.9379} & \textbf{0.7780} & \textbf{0.8199} & \textbf{0.8646} & \textbf{0.9088} &\textbf{0.7688} \\
        CRPO$\times$ & \textbf{0.8489} & \textcolor{gray}{\textbf{0.9015}} & \textcolor{gray}{\textbf{0.9372}} & \textcolor{gray}{\textbf{0.7769}} & \textcolor{gray}{\textbf{0.8176}} & \textcolor{gray}{\textbf{0.8627}} & \textcolor{gray}{\textbf{0.9051}} &\textcolor{gray}{\textbf{0.7655}} \\\hline
    \end{tabular}

    \begin{tabular}{cc}
    \end{tabular}

    \begin{tabular}{l|cccc|cccc}
    \hline
        \multirow{2}{*}{\textbf{Method}} & \multicolumn{4}{c|}{$en\rightarrow ru$} & \multicolumn{4}{c}{$en\rightarrow *$} \\\cline{2-9}
        % & \textbf{K-22} & \textbf{C-22} & \textbf{XC} & \textbf{K-XL} & \textbf{K-22} & \textbf{C-22} & \textbf{XC} & \textbf{K-XL} \\\hline
        & \textbf{KIWI22} & \textbf{COMET22} & \textbf{XCOMET} & \textbf{KIWI-XL} & \textbf{KIWI22} & \textbf{COMET22} & \textbf{XCOMET} & \textbf{KIWI-XL} \\\hline
        QE Ft. & 0.8099 & 0.8758 & 0.9335 & 0.7665 & 0.8265 & 0.8689 & 0.9167 & 0.7511  \\
        Goal Ref. & 0.8297 & - & 0.9241 & 0.7598 & 0.8206 & - & 0.9062 & 0.7488 \\ \hline
        RSO & 0.8373 & 0.8792 & 0.9395 & 0.7488 & 0.8287 & 0.8718 & 0.9210 & 0.7456 \\
        RS-DPO ($\eta$=0.60/0.50) & 0.8292 & 0.8726 & 0.9257 & 0.7581 & 0.8223 & 0.8662 & 0.9097 & 0.7456 \\
        RS-DPO ($\eta$=0.65/0.55) & 0.8302 & 0.8729 & 0.9284 & 0.7618 & 0.8223 & 0.8664 & 0.9117 & 0.7466 \\
        Triplet Dataset & 0.8326 & 0.8759 & 0.9388 & 0.7704 & 0.8247 & 0.8693 & 0.9203 & 0.7537 \\
        MBR-BW & 0.8332 & 0.8776 & 0.9349 & 0.7683 & 0.8262 & 0.8709 & 0.9156 & 0.7531 \\
        MBR-BMW & 0.8329 & 0.8784 & 0.9334 & 0.7670 & 0.8257 & 0.8709 & 0.9177 & 0.7532 \\\hline
        CRPO$+$ & \textbf{0.8398} & \textcolor{gray}{\textbf{0.8815}} & \textcolor{gray}{\textbf{0.9441}} & \textbf{0.7821} & \textbf{0.8320} & \textbf{0.8752} & \textbf{0.9257} &\textbf{0.7674} \\
        CRPO$\times$ & \textcolor{gray}{\textbf{0.8397}} & \textbf{0.8820} & \textbf{0.9442} & \textcolor{gray}{\textbf{0.7818}} & \textcolor{gray}{\textbf{0.8314}} & \textcolor{gray}{\textbf{0.8748}} & \textcolor{gray}{\textbf{0.9252}} &\textcolor{gray}{\textbf{0.7660}} \\\hline
    \end{tabular}

\end{table*}

\begin{table*}[t]
    \centering
    \small
    \setlength\tabcolsep{3.5pt}
    \caption{Experiment results on translation directions of $*\rightarrow en$. The average result over these 5 translation directions are shown in the column of $*\rightarrow en$.}
    \label{tab:toen}
    
    \begin{tabular}{l|cccc|cccc}
    \hline
        \multirow{2}{*}{\textbf{Method}} & \multicolumn{4}{c|}{$zh\rightarrow en$} & \multicolumn{4}{c}{$de\rightarrow en$} \\\cline{2-9}
        % & \textbf{K-22} & \textbf{C-22} & \textbf{XC} & \textbf{K-XL} & \textbf{K-22} & \textbf{C-22} & \textbf{XC} & \textbf{K-XL} \\\hline
        & \textbf{KIWI22} & \textbf{COMET22} & \textbf{XCOMET} & \textbf{KIWI-XL} & \textbf{KIWI22} & \textbf{COMET22} & \textbf{XCOMET} & \textbf{KIWI-XL} \\\hline
        QE Ft. & 0.7759 & 0.8017 & 0.8690 & 0.6721 & 0.8108 & 0.8437 & 0.9702 & 0.7349 \\
        Goal Ref. & 0.7709 & - & 0.8420 & 0.6674 & 0.7874 & - & 0.9476 & 0.6975 \\\hline
        RSO & 0.7868 & 0.8093 & 0.8739 & 0.6871 & 0.8129 & \textcolor{gray}{\textbf{0.8463}} & 0.9710 & 0.7372  \\
        RS-DPO ($\eta$=0.60/0.50) & 0.7771 & 0.8019 & 0.8631 & 0.6741 & 0.8093 & 0.8427 & 0.9683 & 0.7312  \\
        RS-DPO ($\eta$=0.65/0.55) & 0.7790 & 0.8045 & 0.8671 & 0.6771 & 0.8095 & 0.8430 & 0.9677 & 0.7321  \\
        Triplet Dataset & 0.7835 & 0.8079 & 0.8728 & 0.6827 & 0.8119 & 0.8462 & 0.9711 & 0.7372 \\
        MBR-BW & 0.7850 & 0.8099 & 0.8745 & 0.6834 & 0.8110 & 0.8454 & 0.9692 & 0.7330 \\
        MBR-BMW & 0.7821 & 0.8081 & 0.8722 & 0.6805 & 0.8102 & 0.8449 & 0.9697 & 0.7341 \\\hline
        CRPO$+$ & \textcolor{gray}{\textbf{0.7883}} & \textbf{0.8106} & \textbf{0.8780} & \textbf{0.6905} & \textcolor{gray}{\textbf{0.8132}} & \textbf{0.8469} & \textbf{0.9719} & \textbf{0.7398}  \\
        CRPO$\times$ & \textbf{0.7889} & \textcolor{gray}{\textbf{0.8098}} & \textcolor{gray}{\textbf{0.8767}} & \textcolor{gray}{\textbf{0.6898}} & \textbf{0.8141} & 0.8459 & \textcolor{gray}{\textbf{0.9717}} & \textcolor{gray}{\textbf{0.7390}}  \\\hline
    \end{tabular}

    \begin{tabular}{cc}
    \end{tabular}

    \begin{tabular}{l|cccc|cccc}
    \hline
        \multirow{2}{*}{\textbf{Method}} & \multicolumn{4}{c|}{$cs\rightarrow en$} & \multicolumn{4}{c}{$is\rightarrow en$} \\\cline{2-9}
        % & \textbf{K-22} & \textbf{C-22} & \textbf{XC} & \textbf{K-XL} & \textbf{K-22} & \textbf{C-22} & \textbf{XC} & \textbf{K-XL} \\\hline
        & \textbf{KIWI22} & \textbf{COMET22} & \textbf{XCOMET} & \textbf{KIWI-XL} & \textbf{KIWI22} & \textbf{COMET22} & \textbf{XCOMET} & \textbf{KIWI-XL} \\\hline
        QE Ft. & 0.8197 & 0.8614 & 0.9383 & 0.7160 & 0.8108 & 0.8639 & 0.9307 & 0.7413 \\
        Goal Ref. & 0.8209 & - & 0.9325 & 0.7150 & 0.8089 & - & 0.9240 & 0.7379 \\\hline
        RSO & \textcolor{gray}{\textbf{0.8239}} & \textbf{0.8650} & 0.9410 & 0.7210 & 0.8138 & 0.8674 & 0.9326 & 0.7431  \\
        RS-DPO ($\eta$=0.60/0.50) & 0.8179 & 0.8613 & 0.9365 & 0.7122 & 0.8073 & 0.8615 & 0.9244 & 0.7363  \\
        RS-DPO ($\eta$=0.65/0.55) & 0.8199 & 0.8623 & 0.9374 & 0.7147 & 0.8079 & 0.8621 & 0.9245 & 0.7364  \\
        Triplet Dataset & 0.8213 & 0.8633 & 0.9427 & 0.7187 & 0.8130 & 0.8662 & 0.9322 & 0.7412 \\
        MBR-BW & 0.8220 & 0.8633 & 0.9386 & 0.7171 & 0.8123 & 0.8664 & 0.9308 & 0.7401 \\
        MBR-BMW & 0.8219 & 0.8652 & 0.9403 & 0.7165 & 0.8106 & 0.8643 & 0.9275 & 0.7391 \\\hline
        CRPO$+$ & \textbf{0.8240} & \textcolor{gray}{\textbf{0.8644}} & \textbf{0.9442} & \textbf{0.7241} & \textcolor{gray}{\textbf{0.8152}} & \textbf{0.8682} & \textbf{0.9338} & \textcolor{gray}{\textbf{0.7450}}  \\
        CRPO$\times$ & 0.8238 & 0.8633 & \textcolor{gray}{\textbf{0.9422}} & \textcolor{gray}{\textbf{0.7220}} & \textbf{0.8160} & \textbf{0.8682} & \textcolor{gray}{\textbf{0.9366}} & \textbf{0.7457}  \\\hline
    \end{tabular}

    \begin{tabular}{cc}
    \end{tabular}

    \begin{tabular}{l|cccc|cccc}
    \hline
        \multirow{2}{*}{\textbf{Method}} & \multicolumn{4}{c|}{$ru\rightarrow en$} & \multicolumn{4}{c}{$*\rightarrow en$} \\\cline{2-9}
        % & \textbf{K-22} & \textbf{C-22} & \textbf{XC} & \textbf{K-XL} & \textbf{K-22} & \textbf{C-22} & \textbf{XC} & \textbf{K-XL} \\\hline
        & \textbf{KIWI22} & \textbf{COMET22} & \textbf{XCOMET} & \textbf{KIWI-XL} & \textbf{KIWI22} & \textbf{COMET22} & \textbf{XCOMET} & \textbf{KIWI-XL} \\\hline
        QE Ft. & 0.8121 & 0.8480 & 0.9512 & 0.7184 & 0.8059 & 0.8437 & 0.9319 & 0.7165 \\
        Goal Ref. & 0.8074 & - & 0.9403 & 0.7066 & 0.7991 & - & 0.9173 & 0.7048 \\\hline
        RSO & 0.8158 & \textcolor{gray}{\textbf{0.8511}} & 0.9528 & 0.7223 & 0.8106 & \textcolor{gray}{\textbf{0.8478}} & 0.9343 & 0.7217  \\
        RS-DPO ($\eta$=0.60/0.50) & 0.8109 & 0.8489 & 0.9483 & 0.7168 & 0.8045 & 0.8433 & 0.9281 & 0.7141  \\
        RS-DPO ($\eta$=0.65/0.55) & 0.8122 & 0.8493 & 0.9497 & 0.7176 & 0.8057 & 0.8442 & 0.9293 & 0.7156  \\
        Triplet Dataset & 0.8148 & 0.8506 & 0.9538 & 0.7224 & 0.8089 & 0.8468 & 0.9345 & 0.7204 \\
        MBR-BW & 0.8131 & 0.8500 & 0.9482 & 0.7178 & 0.8087 & 0.8470 & 0.9323 & 0.7183\\
        MBR-BMW & 0.8133 & 0.8516 & 0.9500 & 0.7185 & 0.8076 & 0.8468 & 0.9319 & 0.7177\\\hline
        CRPO$+$ & \textbf{0.8172} & \textbf{0.8517} & \textbf{0.9547} & \textbf{0.7259} & \textcolor{gray}{\textbf{0.8116}} & \textbf{0.8484} & \textbf{0.9365} & \textbf{0.7251}  \\
        CRPO$\times$ & \textcolor{gray}{\textbf{0.8171}} & 0.8510 & \textcolor{gray}{\textbf{0.9539}} & \textcolor{gray}{\textbf{0.7247}} & \textbf{0.8120} & 0.8476 & \textcolor{gray}{\textbf{0.9362}} & \textcolor{gray}{\textbf{0.7242}}  \\\hline
    \end{tabular}

\end{table*}

\subsection{Result on Each Translation Direction}
\label{sec:ap_direct_result}

To further evaluate the performance of CRPO, we show the evaluation results on $en\rightarrow *$ in Table~\ref{tab:ento} and $*\rightarrow en$ in Table~\ref{tab:toen}. CRPO achieves the best results on almost all directions which empirically proves the robustness of our method and the importance of leveraging confidence term to measure the policy performance for different translation directions.

\subsection{Non-COMET Metric}
\label{sec:bleurt}
Due to the similar training procedure of COMET metrics, concerns may arise that the results in Section~\ref{sec:exp} and Appendix~\ref{sec:ap_direct_result} could be highly correlated with the reward model - COMET metrics. To address this concern, we also consider BLEURT-20~\cite{sellam2020bleurt} for evaluation, which is a non-COMET and neural-based metric. We compare CRPO with RSO, RS-DPO, MBR Score and Triplet Dataset for both ALMA-7b and NLLB-1.3B models in Table~\ref{tab:bleurt}, where CRPO$+$ and CRPO$\times$ achieve the best score indicating the robustness and high performance of combining confidence and reward terms.

\begin{table}[]
    \centering
    \small
    \setlength\tabcolsep{2.5pt}
    \caption{Evaluation Results based on BLEURT-20 metric for ALMA-7B and NLLB-1.3B.}
    
    \begin{tabular}{l|cc}
        \textbf{Method} & \textbf{ALMA-7B} & \textbf{NLLB-1.3B} \\\hline
        RSO & 0.7451 & 0.7305 \\
        RS-DPO ($\eta$=0.60/0.50) & 0.7401 & 0.7154 \\
        RS-DPO ($\eta$=0.65/0.55) & 0.7404 & 0.7158 \\
        Triplet Dataset & 0.7444 & 0.7212 \\
        MBR-BW & 0.7463 & 0.7281 \\
        MBR-BMW & 0.7460 & 0.7251 \\\hline
        CRPO$+$ & \textbf{0.7497} & \textbf{0.7317} \\
        CRPO$\times$ & \textcolor{gray}{\textbf{0.7490}} & \textcolor{gray}{\textbf{0.7314}}
    \end{tabular}
    \label{tab:bleurt}
\end{table}

\subsection{Visualization of Reward and Confidence}

\begin{figure}[t]
    \centering
    \includegraphics[width=\linewidth]{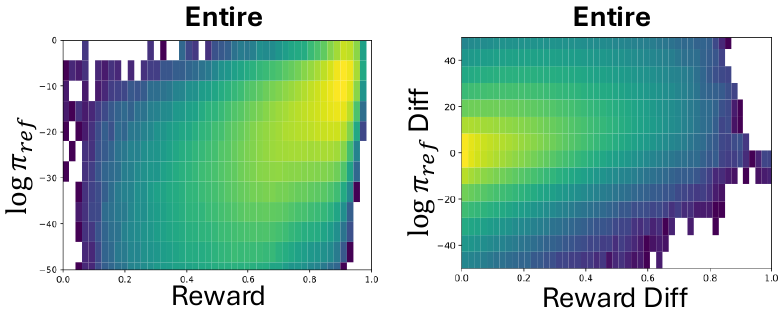}
    \caption{Reward score vs. $\log\pi_{ref}$ for entire dataset.}
    \label{fig:entire}
\end{figure}

\begin{figure*}[t]
    \centering
    \begin{subfigure}{\textwidth}
        \centering
        \includegraphics[width=\textwidth]{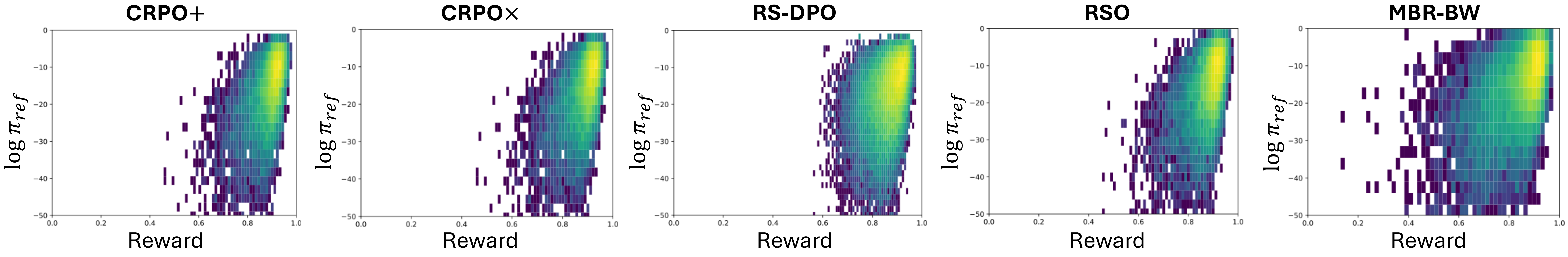}
        \caption{Preferred Sentences.}
        \vspace{0.1cm}
    \end{subfigure}
    \begin{subfigure}{\textwidth}
        \centering
        \includegraphics[width=\textwidth]{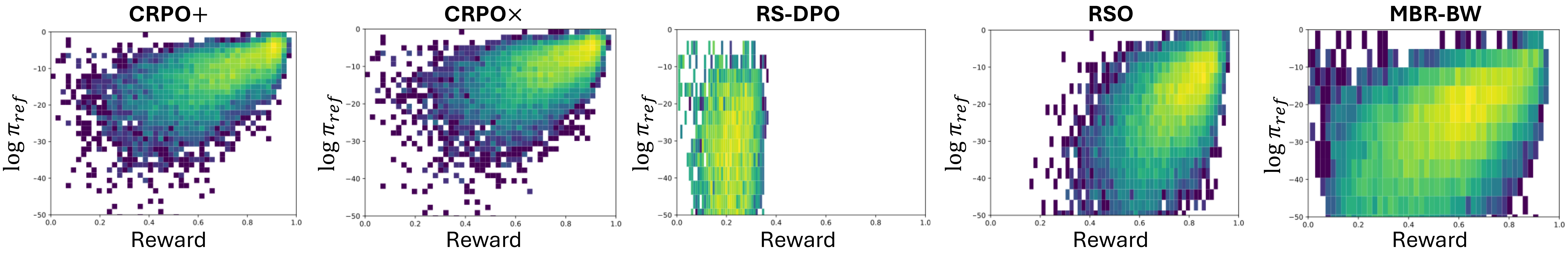}
        \caption{Dis-preferred Sentences.}
        \vspace{0.1cm}
    \end{subfigure}
    \begin{subfigure}{\textwidth}
        \centering
        \includegraphics[width=\textwidth]{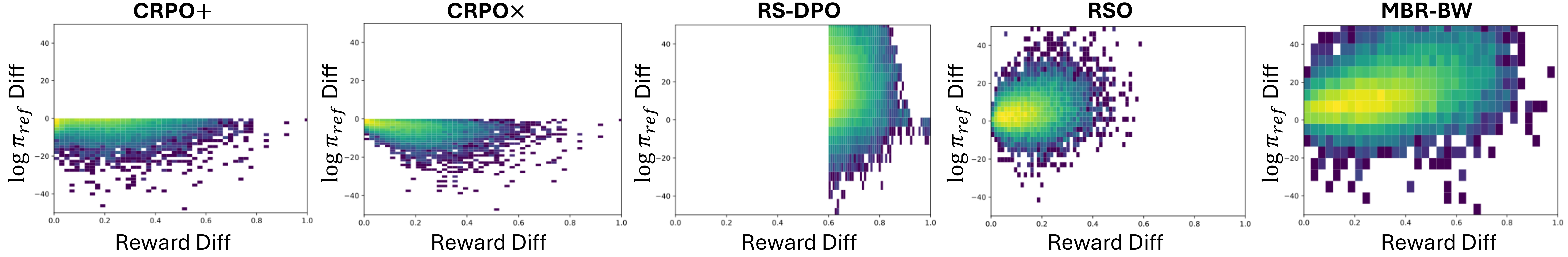}
        \caption{Reward Difference vs. $\log\pi_{ref}$ Difference.}
    \end{subfigure}
    \caption{Distribution of reward scores and $\log\pi_{ref}$.}
    \label{fig:vis_dist_2}
\end{figure*}

To further represent the CRPO strategy, we visualize the correlation between reward scores and $\log\pi_{ref}$ for candidate dataset in Figure~\ref{fig:entire} and for selected dataset in Figure~\ref{fig:vis_dist_2}. 

As the ALMA-7B checkpoint already achieves outstanding performance, Figure~\ref{fig:entire} shows that before fine-tuning the policy has high probability to generate high reward translation sentences and sentence pairs with low reward difference also tend to have low $\log\pi_{ref}$ difference. This also explains that in Figure~\ref{fig:vis_dist_2}, for all data selection methods, sentences with high reward tend to have high generation likelihood. But comparing with RS-DPO, RSO and MBR-BW, dis-preferred sentences selected by CRPO generally get higher value of $\log\pi_{ref}$ which refers to difficult or error predicted translation sentences for the policy. Moreover, as CRPO only selects sentence pairs with positive CR-Score, only sentence pairs with negative $\log\pi_{ref}$ difference are remained in Figure~\ref{fig:vis_dist_2}.

\subsection{Potential Questions.}
\label{sec:app_qa}

We further evaluate CRPO by considering potential questions could be raised from CR-Score and attempt to answer them with existed or novel experiment results.

\textbf{Is it possible to apply CRPO on sentences from extra resources with higher reward?} To answer this question, we compose a new candidate set by mixing Triplet Dataset with our generated candidate sentences from reference policy. We then apply CR-Score to construct preference dataset and fine-tune the policy with DPO. The results are shown in Table~\ref{tab:mix} where mixing Triplet Dataset (CRPO$+^*$ and CRPO$\times^*$) increases the overall performance of CRPO. Note that although fine-tuning Triplet Dataset gets worse evaluation scores, CRPO selects sentences from extra resource only when they could provide useful information for the policy and thus achieves higher performance.

\begin{table}[t]
    \centering
    \small
    \setlength\tabcolsep{2.pt}
    \caption{Average results of CRPO on mixed dataset.}
    \label{tab:mix}
    \begin{tabular}{l|cccc}
    \hline
        \multirow{2}{*}{\textbf{Method}} & \multicolumn{4}{c}{Average} \\\cline{2-5}
         % & \textbf{K-22} & \textbf{C-22} & \textbf{XC} & \textbf{K-XL} \\\hline
         & \textbf{KIWI22} & \textbf{COMET22} & \textbf{XCOMET} & \textbf{KIWI-XL} \\\hline
        CRPO$+$ & 0.8218 & 0.8618 & \textcolor{gray}{\textbf{0.9311}} & \textcolor{gray}{\textbf{0.7462}} \\
        CRPO$\times$ & 0.8217 & 0.8612 & 0.9307 & 0.7451 \\\hline
        Triplet & 0.8168 & 0.8581 & 0.9274 & 0.7371 \\
        CRPO$+^*$ & \textbf{0.8223} & \textcolor{gray}{\textbf{0.8622}} & 0.9299 & 0.7458 \\
        CRPO$\times^*$ & \textcolor{gray}{\textbf{0.8221}} & \textbf{0.8626} & \textbf{0.9319} & \textbf{0.7465} \\\hline
    \end{tabular}
\end{table}

\begin{table}[t]
    \centering
    \small
    \setlength\tabcolsep{1.2pt}
    \caption{Average results of CRPO and Triplet Dataset on CPO.}
    \label{tab:cpo}
    \begin{tabular}{l|cccc}
    \hline
        \multirow{2}{*}{\textbf{Method}} & \multicolumn{4}{c}{Average} \\\cline{2-5}
         % & \textbf{K-22} & \textbf{C-22} & \textbf{XC} & \textbf{K-XL} \\\hline
         & \textbf{KIWI22} & \textbf{COMET22} & \textbf{XCOMET} & \textbf{KIWI-XL} \\\hline
        Triplet (CPO) & 0.8175 & 0.8587 & 0.9284 & 0.7389 \\
        CRPO$+$ (CPO) & \textbf{0.8214} & \textbf{0.8607} & \textbf{0.9306} & \textbf{0.7451} \\
        CRPO$\times$ (CPO) & \textbf{0.8214} & 0.8604 & 0.9302 & 0.7450 \\\hline
    \end{tabular}

    \begin{tabular}{cc}
    \end{tabular}

\end{table}

\textbf{Will Triplet Dataset achieves better result when trained with CPO?} For the reason that reference policy is dropped in CPO, distribution shift problem might be released in CPO and the quality of sentence pair should be more important. To answer the question, we provide additional experiment to fine-tune the policy on CPO, the results of which are shown in Table~\ref{tab:cpo}. CRPO with CPO fine-tuning still achieves better overall performance compared with Triplet Dataset. Moreover, it is interesting to show that although Triplet Dataset achieves better result with CPO compared with DPO, CPO on preference dataset constructed by CR-Score does not achieves better performance compared with DPO. We think the reason is that preference dataset from CRPO already provides enough information to increase the reward that policy could achieve and dropping the KL divergence term in RLHF objective such as CPO would not further improve the performance. The better way to increase fine-tuning result of CRPO is to increase the quality of sentences, such as mixing with Triplet Dataset.

\end{document}